\def\year{2022}\relax
\documentclass[letterpaper]{article} 
\usepackage{aaai22}  
\usepackage{times}  
\usepackage{helvet}  
\usepackage{courier}  
\usepackage[hyphens]{url}  
\usepackage{graphicx} 
\usepackage{natbib}  
\usepackage{caption} 
\DeclareCaptionStyle{ruled}{labelfont=normalfont,labelsep=colon,strut=off} 
\usepackage[nottoc]{tocbibind}

\usepackage{soul}
\usepackage{amsmath}
\usepackage{booktabs}
\usepackage[utf8]{inputenc}
\usepackage{amsfonts}
\usepackage{xspace}
\usepackage{enumitem}
\usepackage{array}
\usepackage{multirow}
\usepackage{subcaption}
\usepackage{tablefootnote}
\usepackage[export]{adjustbox}
\graphicspath{ {Figures/} }
\usepackage[ruled,vlined]{algorithm2e}  
\SetKwInOut{KwInput}{Input} 
\SetKwInOut{KwOutput}{Output} 
\usepackage{xcolor}

\title{Topo2vec: Topography Embedding Using the Fractal Effect}
\author {
Jonathan Kavitzky, \textsuperscript{\rm 1,\rm 2}
Jonathan Zarecki, \textsuperscript{\rm 1,\rm 3}
Idan Brusilovsky, \textsuperscript{\rm 1,\rm 4}
Uriel Singer\textsuperscript{\rm 1,\rm 5}
}
\affiliations {
    \textsuperscript{\rm 1} Penta-AI \\
    \textsuperscript{\rm 2} HUJI: The Hebrew University, Jerusalem, Israel \\
    \textsuperscript{\rm 3} Bar-Ilan University, Ramat Gan, Israel \\
    \textsuperscript{\rm 4} The Open University, Ra'anana, Israel \\
    \textsuperscript{\rm 5} Technion---Israel Institute of Technology, Haifa, Israel \\
    jonathan.kavitzky@mail.huji.ac.il, jonathan.zarecki@gmail.com, brusli1@gmail.com, urielsinger@cs.technion.ac.il
}

\begin{document}

\maketitle

\newcommand{\specialcell}[2][c]{%
  \begin{tabular}[#1]{@{}c@{}}#2\end{tabular}}

\newcommand{\znote}[1]{\textcolor{blue}{$\ll$\textsf{#1 --Zarecki}$\gg$}}
\newcommand{\snote}[1]{\textcolor{red}{$\ll$\textsf{#1 --Singer}$\gg$}} 
\newcommand{\knote}[1]{\textcolor{orange}{$\ll$\textsf{#1 --Kavitz}$\gg$}}
\newcommand{\bnote}[1]{\textcolor{green}{$\ll$\textsf{#1 --Brus}$\gg$}}

\newcommand\footnoteref[1]{\protected@xdef\@thefnmark{\ref{#1}}\@footnotemark}

\newcommand{\ttv}{{topo2vec}\xspace}
\newcommand{\Ttv}{{Topo2vec}\xspace}

\newcommand{\argmin}{\mathop{\mathrm{argmin}}} 

\newcolumntype{L}[1]{>{\raggedright\let\newline\\\arraybackslash\hspace{0pt}}m{#1}}
\newcolumntype{C}[1]{>{\centering\let\newline\\\arraybackslash\hspace{0pt}}m{#1}}
\newcolumntype{R}[1]{>{\raggedleft\let\newline\\\arraybackslash\hspace{0pt}}m{#1}}

\def\EE{{\mathbb E}}
\def\PP{{\mathbb P}}
\begin{abstract}
Recent advances in deep learning have transformed many fields by introducing generic embedding spaces, capable of achieving great predictive performance with minimal labeling effort. The geology field has not yet met such success.
In this work, we introduce an extension for self-supervised learning techniques tailored for exploiting the fractal-effect in remote-sensing images.
The fractal-effect assumes that the same structures (for example rivers, peaks and saddles) will appear in all scales.
We demonstrate our method's effectiveness on elevation data, we also use the effect in inference.
We perform an extensive analysis on several classification tasks and emphasize its effectiveness in detecting the same class on different scales.
To the best of our knowledge, it is the first attempt to build a generic representation for topographic images.
\end{abstract}

\section{Introduction}
\label{sec:intro}
From ancient times, the topographic structure of land was a key aspect in many decisions.
The topography of the land is provably correlated with many other tasks: land-use \cite[]{sheikh2014relationship}, soil mapping \cite[]{scull2003predictive}, soil salinity \cite[]{Vermeulen:2017:Geoderma}, landslides \cite[]{prakash2020mapping} water floods \cite[]{cnn4flood2019}, avalanches \cite[]{JAEDICKE201431} and high solar-energy locations \cite[]{HEO2020114588}. 
The techniques for perceiving, collecting, and understanding topography has changed significantly in recent years
and today, geographic information systems (GIS) are built on many classical and data-driven algorithms.

\begin{figure*}[ht]
    \centering
    \includegraphics[width=0.9\linewidth]{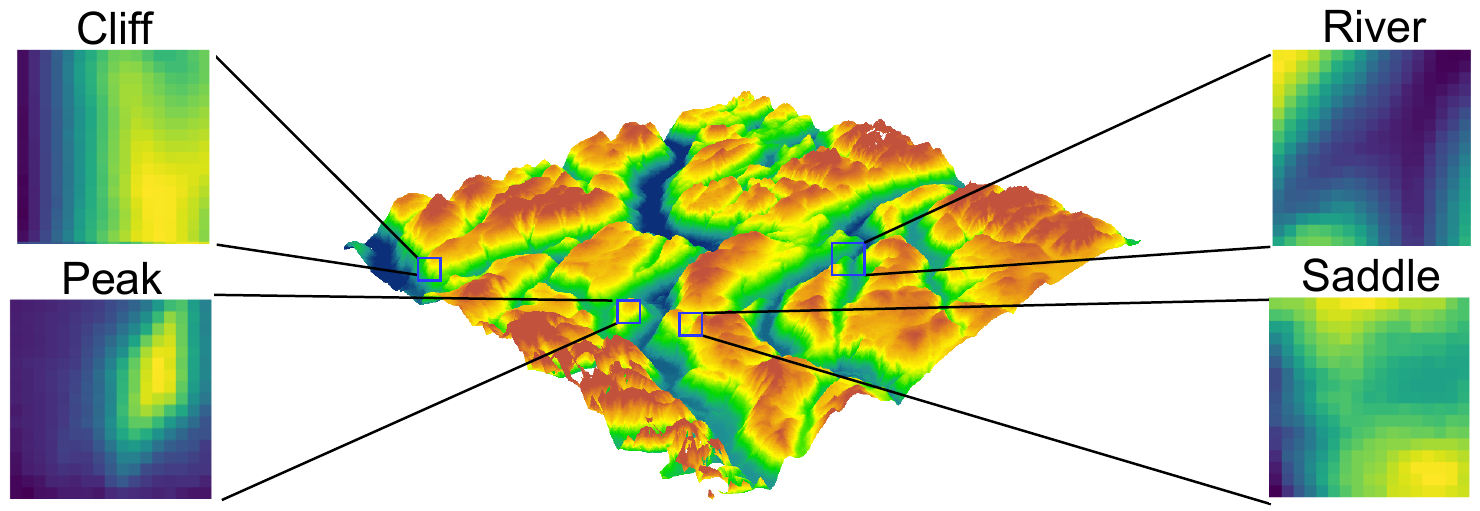}
    \caption{DTM Topography image \footnotemark. Each pixel's value represents its elevation above sea level, and all the values are normalized for each tile. 
    }
    \label{fig:topography_image}
\end{figure*}

As in many fields in recent years, deep learning has also begun to transform the field of topography and GIS in general, with works from automatic road-mapping \cite[]{Mttyus2017DeepRoadMapperER, Zhang2018RoadEB} to elevation map super-resolution \cite[]{Yue2015FusionOM}.
Deep learning techniques have a particularly interesting property with respect to classical ML models: In addition to superior predictive accuracy, they naturally build a latent embedding space in which every example can be mapped to a latent vector.
These embedding spaces have received much attention in self-supervised contrastive learning methods \cite[]{He2020MomentumCF, Chen2020ASF}, where an intrinsic property in the data is used to train the model instead of labels gathered by human annotators. Self-supervised methods have been used to achieve close-to-SOTA results on benchmarks such as ImageNet \cite[]{Russakovsky2015ImageNetLS}, iNaturalist \cite[]{Horn2018TheIS}, Birdsnap \cite[]{Berg2014BirdsnapLF}, and many more \cite[]{Nilsback2008AutomatedFC, FeiFei2004LearningGV}.
These methods usually exploit specific invariants in the data, such as invariance to rotation, color jitter, cropping, and more.

Topography data exhibits an interesting property which we call the \emph{fractal-effect}. 
This effect implies that the same location will have the same topographic patterns  when viewed in different radii/scales. Topographic objects such as peaks and saddles will appear in every observed scale.
This property also implies in practice that any embedding space for topographic data should be highly scale-dependent, and choosing a scale is highly relevant to how the data would be represented;
that is, the same location with different scales should have very different embeddings since they represent different things.

In this work, we present a new self-supervised technique tailored for topographic images, we exploit the fact that topography images \cite[]{Pelletier1997WhyIT} are built as fractals and train a neural network capable of embedding the images in a useful latent embedding space. 
We use the fractal-effect in inference and prove our embedding space is effective in many scales.
Our main contributions are as follows: 
\begin{enumerate}
\item We introduce a self-supervised training technique for deep neural networks tailored for topography data and use this technique to train a model capable of embedding a topographic image into a useful embedding space.
\item We empirically evaluate the model and technique on classification benchmarks we have built for topographic images, and present qualitative results expressing the fractal-effect in inference.
\item We build a baseline comparison for future work in the field of topography ML. This includes: data, architecture, tasks, and metrics.
\end{enumerate}

\section{Exploiting the Fractal-Effect for Self-Supervision}
\label{sec:framework}
\footnotetext{\url{gisgeography.com}}

In this section, we will first introduce the fractal-effect in topography images, afterward, we will discuss the data used in this paper, and then we will define our self-supervision technique exploiting the fractal-effect.

\subsection{Fractal Effect}
\label{ssec:fractaleffect}

\begin{figure}[ht]
    \centering
    \includegraphics[width=0.45\textwidth]{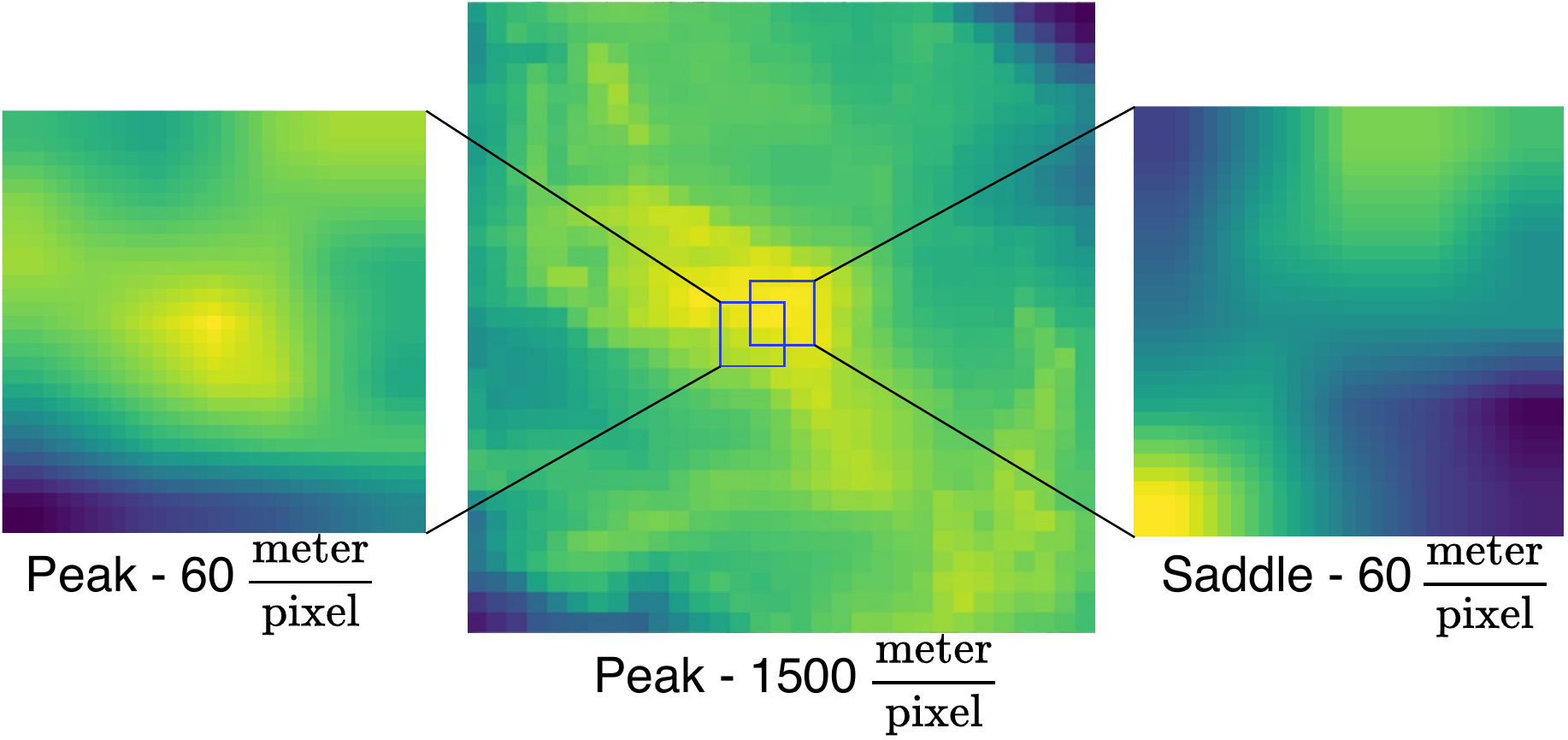}
    \caption{2 topographic images that were taken from two close points.
    In a small scale, the two images look very differently and are classified as a peak and a saddle, but in a bigger scale, both locations are of a peak. Different scales imply different topography. 
    Taken in: $46^\circ51'34.2"N$ $7^\circ31'31.8"E$.}
    \label{fig:fractal-effect_multiscale}
\end{figure}

As discussed in the introduction, topography data exerts many interesting properties.
One of the main properties of the topographic data is scale semi-invariance/semi-fractal: the data can be viewed as topographic in many resolutions and is somewhat similar in those different resolutions \cite[]{jin2017definition}, we call this property the \emph{fractal-effect}.
For example, as seen in Figure \ref{fig:fractal-effect_multiscale}, peaks appear in both very high and low zoom levels of the topography.

As a result of this effect, and in contrast to most computer-vision tasks, our interest objects will certainly appear in multiple scales and any model working in this domain must exert a high level of scale-dependence. 
Now, in order to support multiple scales, one can change the image resolution so that it will be observed at a different scale. See Figure \ref{fig:fractal-effect_multiscale} for a visual illustration. 

However, it is important to remember that the fractal-effect implies that each image in the data contains objects in multiple scales. 
As before performing any experiments one should decide which scale (or multiple scales) the data was sampled at.

\subsection{Data}
\subsubsection{Digital terrain model (DTM)}
In order to create the topography images, we used the DTM elevation data from \cite{ALOS}.
Every topography image is a 2D array of pixels, each representing the ground surface elevation above sea level. This database has a spatial resolution of 30 meters per pixel, and elevation accuracy of $\pm$10 meters. In our experiments
we used Europe as an area of interest.

\subsubsection{OpenStreetMap (OSM)}
For precise locations of objects of different classes, used later for evaluation, we used OpenStreetMap \cite[]{haklay2008openstreetmap}.
Each location is represented by a pair of geographic coordinates (longitude, latitude), which in turn will be converted to a topography image. We used geo-locations of different topographic classes (peaks, rivers and saddles), and non-topographic classes (aerialway stations) all sampled evenly at random out of all the OSM known tags.

\subsection{Topo2Vec \label{ssec:training_technique}}

Let $loc$ be a location on earth:
$x_r^s$ is the image of the area with radius $r$  around $loc$ (as presented in Figure \ref{fig:topography_image}),
N is the number of pixels from the image's central to the edge.
The spatial resolution in $\frac{\text{meter}}{\text{pixel}}$ of $x_r^s$ is defined as: 
\[s = r / N \frac{\text{meter}}{\text{pixel}}\]

One can sample the image at different spatial resolutions $x_r^{s_1}, x_r^{s_2}, \dots, x_r^{s_j}$, while covering the same area. 
However, due to the fractal-effect,  different topographic objects in the observed area will become more apparent in different resolutions, as seen in Figure \ref{fig:fractal-effect_multiscale}.

Taking the fractal-effect into consideration,
we define $f$ as a function that encodes $x_r^{s}$ into an embedding space. Our goal is that $f(x_r^{s})$ will represent the topographic information present around $loc$ with radius $r$ and a resolution of $s$. 
By learning a second function, $g$, that decodes $f(x_r^{s})$ to an image with a different spatial resolution $s'$, we are able to leverage the fractal-effect and learn a better topographic representation.
This is achieved by maximizing:
$$\PP(x_r^{s'}|g(f(x_r^{s})))$$

\noindent
Where $s'=k\cdot s$ \scalebox{0.5}{$(k \geq 1)$} can be thought as the \emph{fractal-factor}.

Maximizing the above expression is done by minimizing the $L^p$ distance:
$$\min_{\theta_{f}} L_p = \|x_r^{s'}, g(f(x_r^{s}))\|$$

Training using this loss, we obtain two functions $f$ and $g$. While $g$ decodes a given embedding vector into a higher resolution image, meaning it learned the fractal-effect around $loc$, $f$ (from now on will be called \emph{\ttv}) is a self-supervised, fractal-effect aware encoder that takes as input a topographic image and embeds it into a embedding space.

\begin{algorithm}
	\KwInput{$Locations$ - list of different locations \\
	        $Scales$ - list of relevant scales \\
	        $k$ - desired fractal-factor \\
	        $lr_1, lr_2$ - learning rates}
	\KwOutput{$f$ - learned encoder \\
            $g$ - learned decoder \\
            $d$ - learned discriminator}
	\nl	\While{f,g,d not converged}{
	\nl	    \For{$batch$ in $Locations$}{
	\nl	        \For{$s \in Scales$}{
	\nl             $X$ = toImageDataset($batch$, $s$)\;
	\nl             $Y$ = toImageDataset($batch$, $k \cdot s$)\;
	\nl             $\hat{Y} = g(f(X))$\;
	\nl             $L_{p} = \sum_{i} ||y_i, \hat{y_i}||_p$\; 
	\nl             $L^{cGAN}_G = BCE(d([X, \hat{Y}]), \bar{1})$\;
	\nl             $f {\leftarrow} f-lr_1 \cdot \nabla_f({\lambda_1*L_{p} + \lambda_2*L^{cGAN}_G})$\;
	\nl             $g {\leftarrow} g-lr_1 \cdot \nabla_g({\lambda_1*L_{p} + \lambda_2*L^{cGAN}_G})$\;
	\nl             $L^{cGAN}_D = BCE(d([X, Y]), \bar{1}) + BCE(d([X, \hat{Y}]), \bar{0})$\;
	\nl             $d {\leftarrow} d-lr_2 \cdot \nabla_d({L^{cGAN}_D})$\;
	}}}             
	\nl	return $f, g, d$ \;
\caption{\Ttv training procedure}
\label{alg:topo2vec}
\end{algorithm}

Intuitively, if $loc$ expresses a fractal-effect that one peak becomes many small peaks in higher resolution. Then, if $g$ has managed to predict this fractal-pattern, it has learned the topographic information in $loc$.
A problem that arises from this training procedure is that while $g$ attempts to predict the fractal-pattern, it is not guaranteed that it will predict the exact fractal-effect of $loc$, but rather the distribution of patterns possible in that area.
In order to address this issue, we propose an adversarial loss on top of the current loss to enable the network to predict a sample from the fractal-pattern distribution rather than the distribution's average. 
The discriminator $d$ takes as input fake pairs, $(x_r^s, g(f(x_r^{s})))$, and true pairs, $(x_r^s, x_r^{s'})$ and tries to distinguish between them, while $g\circ f$ tries to fool him. Formally, the adversarial loss is:
\begin{equation*}
\begin{split}
\min_{\theta_{f}\theta_{g}}\max_{\theta_{d}} L_G^{cGAN} = \EE_{s \sim (x_r^s, x_r^{s'})}log(d(s)) \\
+ \EE_{s \sim (x_r^s, g(f(x_r^{s})))}log(1-d(s))
\end{split}
\end{equation*}

While the final optimization loss of \ttv is:
$$\min_{\theta_{f}\theta_{g}}\max_{\theta_{d}} \lambda_1*L_{p} + \lambda_2*L_G^{cGAN}$$

Where $\lambda_1=100$ and $\lambda_2=1$, and $L_G^{cGAN}$ is calculated using binary cross-entropy ($BCE$).
Alg \ref{alg:topo2vec} presents the training procedure of the adversarial version of \ttv, while non-adversarial \ttv trains the same but with $\lambda_1=1$ and $\lambda_2=0$.

\subsection{Implementation Details}
The \ttv architecture consists of down-sample layers and then up-sample layers, similar to the U-net architecture \cite{ronneberger2015u}, but without the skip connections. We take the representation between the down and up-sample layers as the latent representation. The reason we deleted the skip connections is to constrain the latent representation to hold the entire information needed for decoding.
In order to allow learning $s'>s$ we added additional up-sample layers (one for each factor of $2$). We used L1 distance as the loss with a learning rate of $lr_1=0.0002$ (for the adversarial version $g$ had the learning rate of $lr_2=0.0016$). As for the discriminator, we trained a 5-layer CNN with three fully-connected layers at the end.

\subsubsection{Encoder}
The full architecture of the encoder $f$ is as follows:
\begin{enumerate}
    \item The input is a 1x17x17 image representing the topography of a specific coordinate in a given scale.
	\item The input image enters two convolutional layers, each includes a convolution block (8 kernels of size 3x3), BatchNorm \cite{ioffe2015batch}, and ReLu \cite{nair2010rectified}. The image is resized to 16x16 (due to padding), with 8 channels, resulting in an 8x16x16 output.
	\item Four down-sample layers, each includes a max-pooling layer (kernel size of 2x2) followed by two convolutional layers as before, each time doubling the number of filters. After four down-sample layers, we end up with a 128x1x1 image.
\end{enumerate}

\noindent
The 128x1x1 image is flattened into a 128 vector, representing the latent vector of the image.

\subsubsection{Decoder}
The full architecture of the decoder $g$ is as follows:
\begin{enumerate}
    \item The input is a 128x1x1 image representing the last output of $f$.
	\item Four up-sample layers, each includes an up-sampling layer (scale factor of 2 in bilinear mode), followed by two convolutional layers as before, each time decreasing the number of filters by 2. After 4 up-sample layers, we end up with an 8x16x16 image.
	\item In \ttv-1 we finish with a convolutional block that maps the 8x16x16 image to the final 1x16x16 output image.
	\item In \ttv-4 we add two additional up-sample layers that resolve with a 2x64x64 image. We finish with a convolutional block that maps the 2x64x64 image to the final 1x64x64 output image.
\end{enumerate}

\subsubsection{Discriminator}
For the \ttv-adv version, $f$ and $g$ together represent the generator, while $d$ represent the discriminator. The full architecture of the discriminator $d$ is as follows:
\begin{enumerate}
    \item The input is a 2x64x64 image representing in its two channels the true or fake pair of images.
	\item Five layers of convolutional blocks (For the first four: kernel size of 4, stride 2, and padding 1. For the Fifth: kernel size of 4, stride 1, and no padding), BatchNorm, and ReLu. The first starts with 8 filters and each time doubling, ending up with a 128x1x1 image.
	\item Flattening the image to a 128 size vector.
	\item Three fully connected layers: $128\rightarrow64$, $64\rightarrow32$, $32\rightarrow1$, with a ReLu activation after the first two.
	\item Sigmoid activation resolving with a final probability neuron.
\end{enumerate}

\section{Experiments}
\label{sec:experiments}
We test our framework on several topographic classes gathered from OpenStreetMap (OSM), the leading geodata repository, in both few-shot and many-shot settings.
Lastly, we’ll show an experiment to show how we generalize beyond our initially trained scale.

\subsection{Experimental Methodology \label{ssec:exp_method}}

\subsubsection{Datasets}
\begin{figure}[ht]
    \centering
    \includegraphics[width=0.4\textwidth]{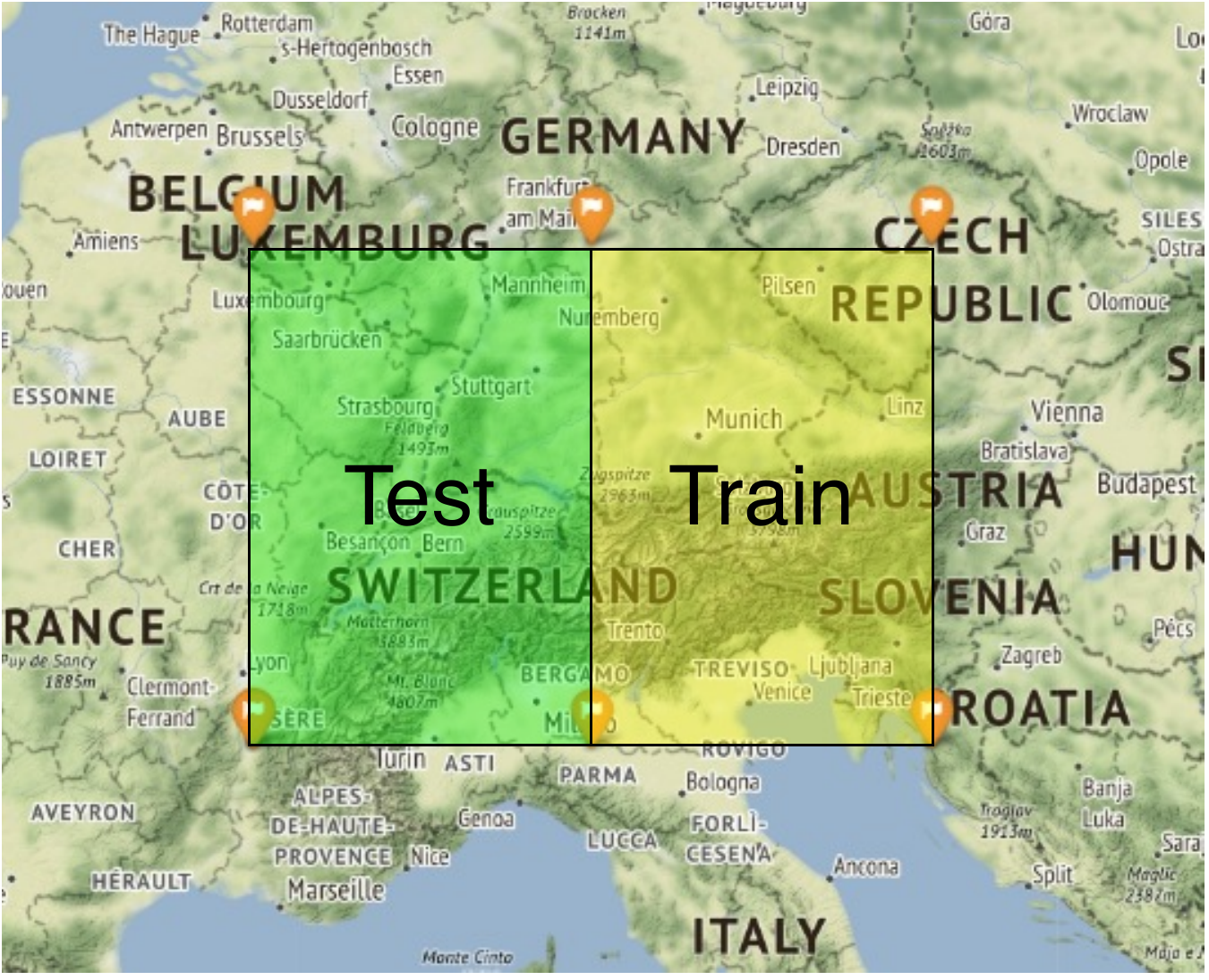}
    \caption{Geographic train/test split. 
    }
    \label{fig:bbox}
\end{figure}

In order to guarantee a fair evaluation procedure, we geographically split the DTM into two different areas of size 25 $degree^2$ each, one for train and one for test (Figure \ref{fig:bbox}).
The polygon from which we sampled coordinates for the train-set is:
$$POLYGON ((10\;\;50,\;10\;\;45,\;15\;\;45,\;15\;\;50,\;10\;\;50))$$
The polygon from which we sampled coordinates for the test set is:
$$POLYGON ((5\;\;45,\;5\;\;50,\;10\;\;50,\;10\;\;45,\;5\;\;45))$$

Each sample was built by extracting topographic data on an image of radius $r$ around a geographic point (as presented in Figure \ref{fig:topography_image}), and resizing it to the appropriate size (17 pixels).

\subsubsection{Baselines}

\begin{itemize}
\item \textbf{ID}: A simple flatten operation on the given image resulting in a vector that is used as the feature representation.

\item \textbf{CNN}: A convolutional network that is trained on the classification of geographic entities from OSM (fully-supervised learning). 
The CNN is made of 2 convolutional layers and 2 fully-connected layers.
The model is trained on examples of peaks, rivers, cliffs, and saddles - the same classes we later examine in Exp. 1, as those are the only "basic" topographic classes with an appropriate amount of data in the OSM for deep networks training, this results in 20,000 training examples. We use the last hidden layer as the feature representation. 

\item \textbf{SauMoCo} \cite[]{saumoco_2020}: An extension of the contrastive method MoCo \cite[]{He2020MomentumCF}. This method uses an additional spatial augmentation technique of sampling a geographically close image. We use this method instead of classic MoCo as other augmentations employed by contrastive methods are irrelevant for topography data, this includes color jitter, random conversion to greyscale, and more. This method replaces using Tile2Vec \cite[]{Jean2019Tile2VecUR} as it clearly outperforms it in spatial image representation. 

\end{itemize}

We additionally compare to several variations of our method:
\begin{itemize}
\item \textbf{\Ttv-1}: \Ttv with fractal-factor $k=1$ (meaning no fractal learning). This variation will help us compare the effectiveness of exploiting the fractal-effect.
\item \textbf{\Ttv-4}: \Ttv with fractal-factor $k=4$.
\item \textbf{\Ttv-adv}: An adversarial version of \ttv network as explained last section with fractal-factor $k=4$.
\end{itemize}

\noindent
All self-supervised methods were trained with 100,000 images. All input images to all methods are the size of $17\times17$\footnote{GitHub repository with all code, baselines, data, and experiments: \url{https://github.com/urielsinger/topo2vec}}.
 
\subsection{Exp. 1: Classification on Topographic Classes}
\label{ssec:exp1}

In this experiment, we evaluate a model trained by our framework’s ability to serve as a pre-training model for topographic tasks with a sufficient amount of data. We test our model individually on 4 topographic classes (rivers\footnote{\url{wiki.openstreetmap.org/wiki/Rivers}}, peaks\footnote{\url{wiki.openstreetmap.org/wiki/Peaks}}, saddles\footnote{\url{wiki.openstreetmap.org/wiki/Tag:natural=saddle }} and cliffs\footnote{\url{wiki.openstreetmap.org/wiki/Tag:natural=cliff}}).

\subsubsection{Finding the class' OSM scale}
A given topographic class is present at many scales, but in OSM, not all scales were collected. 
We'll first devise an experiment to determine what scale the OSM community sampled at:


Let $C=\{c \mid c \in \text{coordinates}\}$ be a coordinate dataset for a given class, constructed such that half of the coordinates are sampled at the location of the class ($y=1$) and half are sampled at random ($y=0$).
Given a radius $r$, we define $X_r^C=\{x^i_r \mid x^i_r \in \mathbb{R}^{17 \times 17} \}$ as the image dataset built from $C$ where each image is built around a coordinate with radius $r$. For each $r$ the spatial resolution changes.

Now, to find the sampled resolution/scale $s$ of set $C$ we define the following experiment:
\begin{enumerate}
    \item Split $X_r^C$ of size 1,000 to train and validation sets (80\%, 20\%).
    \item For each $r \in R$, train a CNN on the available training data and calculate accuracy on the validation set. 
    This CNN has a similar architecture to the CNN baseline.
    \item Repeat this process for 10 different random seeds
    \item The sampled scale is the one with maximum average accuracy on the validation set.
\end{enumerate}

\begin{figure}[t]
    \center\setlength\tabcolsep{3.0pt}\small
    \includegraphics[width=0.5\textwidth]{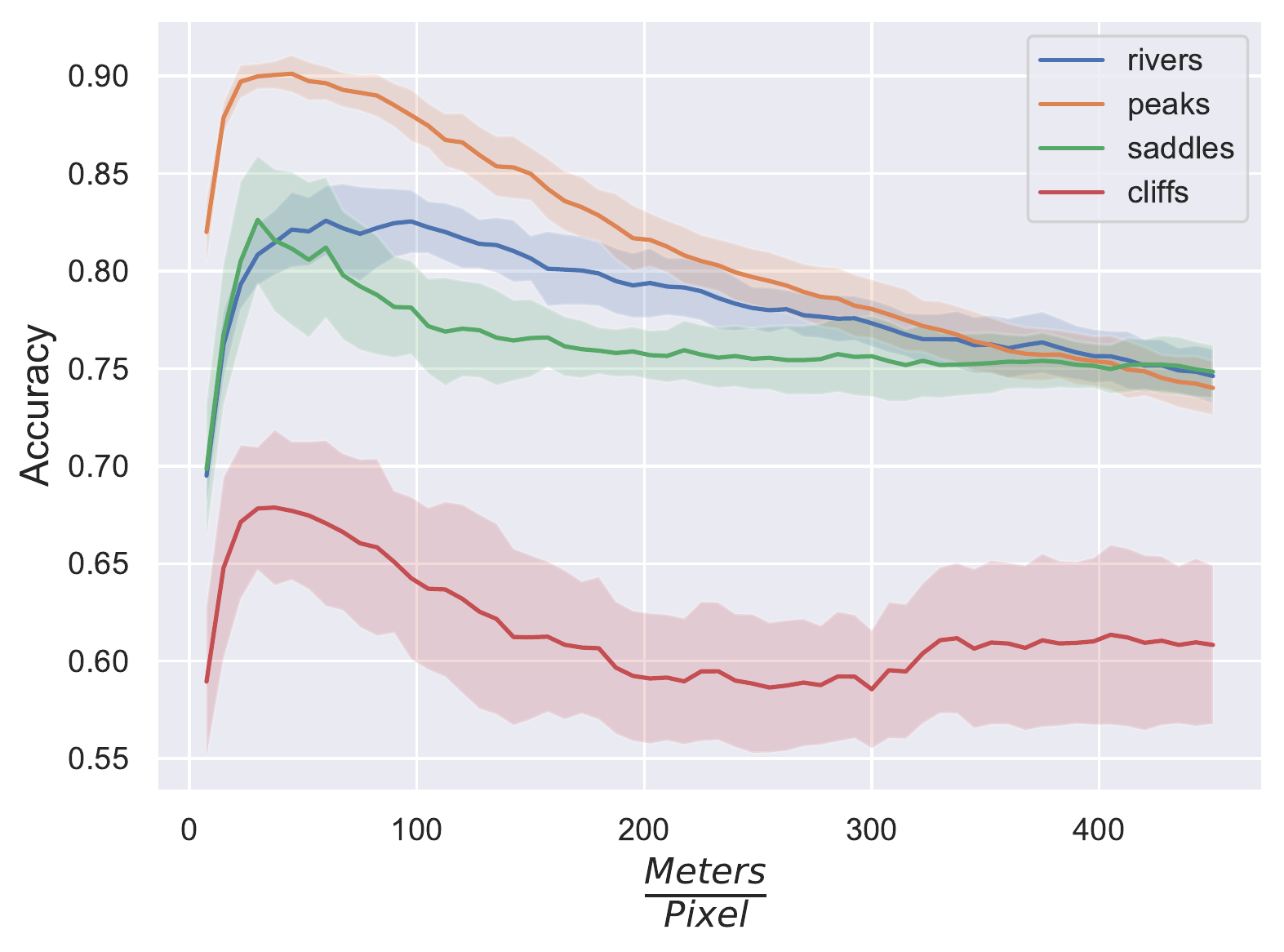}
    \caption{Accuracy-per-scale of the topography classes.}
    \label{fig:scale_selection}
\end{figure}

We can expect the model to have maximum accuracy on the scale of the class' sampled scale, the experiment is repeated for all four classes.
The results of the experiment classes are presented in Figure \ref{fig:scale_selection}.
We observe that all classes peak around the resolution of 30 meters/pixel. 
All subsequent experiments will train the models in this optimal scale.

\subsubsection{Methodology \& Results}
We compared the topo2vec variants with baselines trained on the optimal scale presented above. These models are pre-trained as discussed in \ref{ssec:exp_method}.

For each topographic class and model, we train an SVM built on the model's embeddings with a training set of size 1000 and evaluated on a test set of size 200, both equally distributed between the positive and negative examples.
We repeat each experiment 10 times with random seeds and report average accuracies and standard deviations in Table \ref{tab:exp1_res}. 

\begin{table}[ht]
\center\setlength\tabcolsep{3.0pt}\small
\begin{tabular}{lllll}
\toprule
{} &                    peaks &                   rivers &                   cliffs &                  saddles \\
\midrule
ID       &           $92.4 \pm 0.5$ &           $75.9 \pm 1.8$ &           $74.4 \pm 2.8$ &           $87.5 \pm 1.2$ \\
CNN      &           $92.3 \pm 0.5$ &           $76.4 \pm 0.7$ &           $72.2 \pm 1.8$ &           $84.1 \pm 1.3$ \\
SauMoCo  &           $86.9 \pm 2.3$ &           $81.7 \pm 2.2$ &           $66.2 \pm 2.5$ &           $67.3 \pm 2.3$ \\
\ttv-1   &  $\mathbf{93.5 \pm 0.7}$ &           $81.3 \pm 2.2$ &           $71.4 \pm 1.7$ &           $80.5 \pm 2.3$ \\
\ttv-4   &           $93.1 \pm 0.9$ &           $80.9 \pm 1.7$ &           $75.8 \pm 2.1$ &  $\mathbf{89.0 \pm 1.3}$ \\
\ttv-adv &           $92.8 \pm 0.3$ &  $\mathbf{82.6 \pm 1.2}$ &  $\mathbf{78.1 \pm 2.8}$ &           $88.7 \pm 2.2$ \\
\bottomrule
\end{tabular}
\caption{Experiment 1 results. accuracy $\pm$ standard-deviation}
\label{tab:exp1_res}
\end{table}

We see that our model outperforms other baselines on all classes, even when the baselines are trained on the optimal scale.
It is important to notice that although CNN is a supervised baseline which had access to more labeled data, \ttv outperformed it over all classes. Furthermore, \ttv-4 outperforms \ttv-1 on 2 classes and is comparable on the other two. This shows the power the fractal-effect has in the self-supervised learning procedure. \Ttv-adv outperforms \ttv-4 on two classes and shows the same results on the other two. This indicates that the adversarial loss affects the latent representation learning, although not significantly.
SauMoCo does not perform well in this experiment, we hypothesize this is because several of its base assumptions does not hold in topography data. It assumes that close locations should have the same representation (similar to Tile2Vec \cite[]{Jean2019Tile2VecUR}), and in topography data, where the fractal-effect is present, even the same location can have multiple classes when viewed in different scales. While SauMoCo did not deal with this fact, it is the key consideration of our method.


\subsection{Exp. 2: Classification on Topography-correlated Classes}
\label{ssec:exp2}
We evaluate our method on several topography-correlated classes, where we want to see if the embedding space holds information useful even for classes with only a mild topographic signature.

We test on four classes: aerialway station\footnote{\url{wiki.openstreetmap.org/wiki/Key:aerialway}}, alpine hut\footnote{\url{wiki.openstreetmap.org/wiki/Tag:tourism=alpine_hut}}, waterfall\footnote{\url{wiki.openstreetmap.org/wiki/Waterfalls}} and sinkhole\footnote{\url{wiki.openstreetmap.org/wiki/Tag:natural=sinkhole}}.
Similar to Exp. 1, we train an SVM built on the model's embeddings. We sample a training set of size 400 and a test set of size 100, both equally distributed between the positive and negative examples.
We repeat each experiment 10 times with random seeds.
The average accuracies and standard deviations are summarised in Table \ref{tab:ex2_res}. 


\begin{table}[ht]
\center\setlength\tabcolsep{2.2pt}\small
\begin{tabular}{lllll}
\toprule
{} &       aerialway stations &              alpine huts &                waterfall &                sinkholes \\
\midrule
ID       &           $63.9 \pm 5.5$ &           $59.2 \pm 3.6$ &           $69.2 \pm 2.6$ &           $62.5 \pm 2.4$ \\
CNN      &           $75.5 \pm 2.1$ &           $73.3 \pm 1.6$ &           $69.7 \pm 2.4$ &           $55.3 \pm 2.9$ \\
SauMoCo  &           $69.7 \pm 3.6$ &           $73.3 \pm 2.8$ &           $55.3 \pm 3.1$ &           $56.5 \pm 4.3$ \\
\ttv-1   &           $79.4 \pm 1.5$ &           $70.9 \pm 2.7$ &  $\mathbf{77.9 \pm 2.2}$ &           $60.1 \pm 2.1$ \\
\ttv-4   &           $82.1 \pm 2.0$ &  $\mathbf{75.7 \pm 1.4}$ &           $73.3 \pm 1.6$ &  $\mathbf{63.2 \pm 2.5}$ \\
\ttv-adv &  $\mathbf{82.8 \pm 1.8}$ &           $75.5 \pm 1.3$ &           $74.8 \pm 1.6$ &           $62.9 \pm 2.4$ \\
\bottomrule
\end{tabular}
\caption{Experiment 2 results. accuracy $\pm$ standard-deviation}
\label{tab:ex2_res}
\end{table}

We can see how \ttv significantly outperforms the other baselines on this task. This result emphasizes, wherein the previous experiment all methods were able to learn the topographic classes, it is much harder to predict classes with a less emphasized topographic signature. 
\ttv-4 outperforms \ttv-1 over all classes except waterfall, \ttv-adv show comparable results to \ttv-4. 
The poor performance in the waterfall class can be explained as waterfall do not exhibit fractal properties (no peak inside a waterfall).

This experiment shows that our modifications for topographic data are useful for learning a more powerful representation, as intended. The exploitation of the fractal-effect improves the generality of topography representation that was learned.
Moreover, this suggests that topography data holds additional, uncorrelated information that can be used in addition to normal GIS and tabular features to further improve ML models working in the GIS domain.

\subsection{Exp. 3: Fractal-Effect Visualization}
\label{ssec:fractalvis}

Throughout this work, we repeatedly presented the importance of the fractal-effect and how we might find entirely different objects when looking at different scales.
In this section, we will present several qualitative results of \ttv-4 exhibiting this important property.

\begin{figure}
    \centering
    \subfloat[Empty ground]
        {{\includegraphics[width=0.25\textwidth]
    	{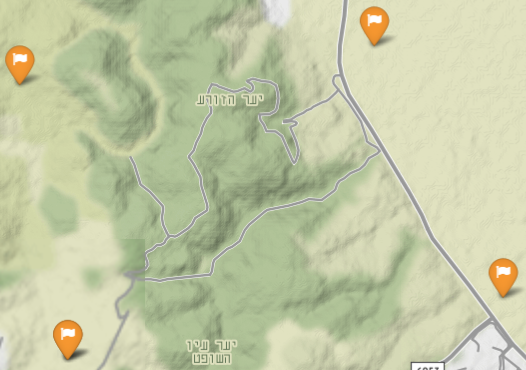} }}
    \subfloat[Low scale - 10 m/pixel]
    	{{\includegraphics[width=0.25\textwidth]
    	{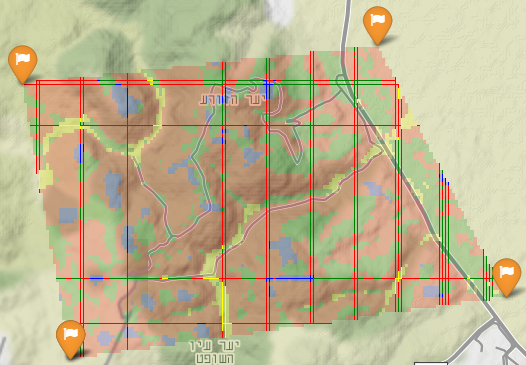} }}
    \quad
    \subfloat[Medium scale - 30 m/pixel]
        {{\includegraphics[width=0.25\textwidth]
    	{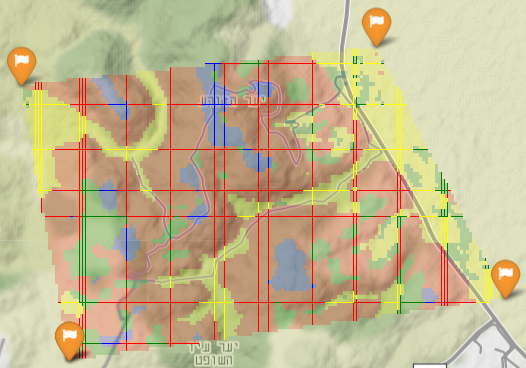} }}
    \subfloat[High scale - 60 m/pixel]
        {{\includegraphics[width=0.25\textwidth]{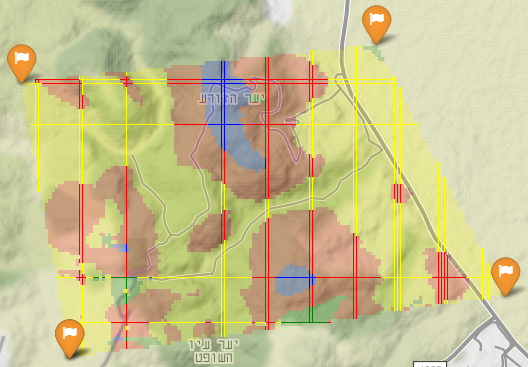} }}
    
    \caption{Qualitative results of our model's predictions in different scales. (blue: peaks, green: saddles, red: cliffs, yellow: rivers) This polygon is around $32^\circ37'02.8"N 35^\circ07'27.4"E.$}
    \label{fig:qual_diff_scales}
\end{figure}

For each point in the polygon, we build an image around it, which is then classified using our model. This process is repeated for different radii and scales.
In Figure \ref{fig:qual_diff_scales} we can see our method detecting peaks, saddles, rivers, and cliffs in multiple scales when images in different zoom levels are presented. 
This is a unique property of topography data and we see our model performs well in this setting.

\subsection{Exp. 4: Topography Retrieval}
\label{ssec:knnvis}
\begin{figure*}[]
    \centering
    \includegraphics[width=1\textwidth]{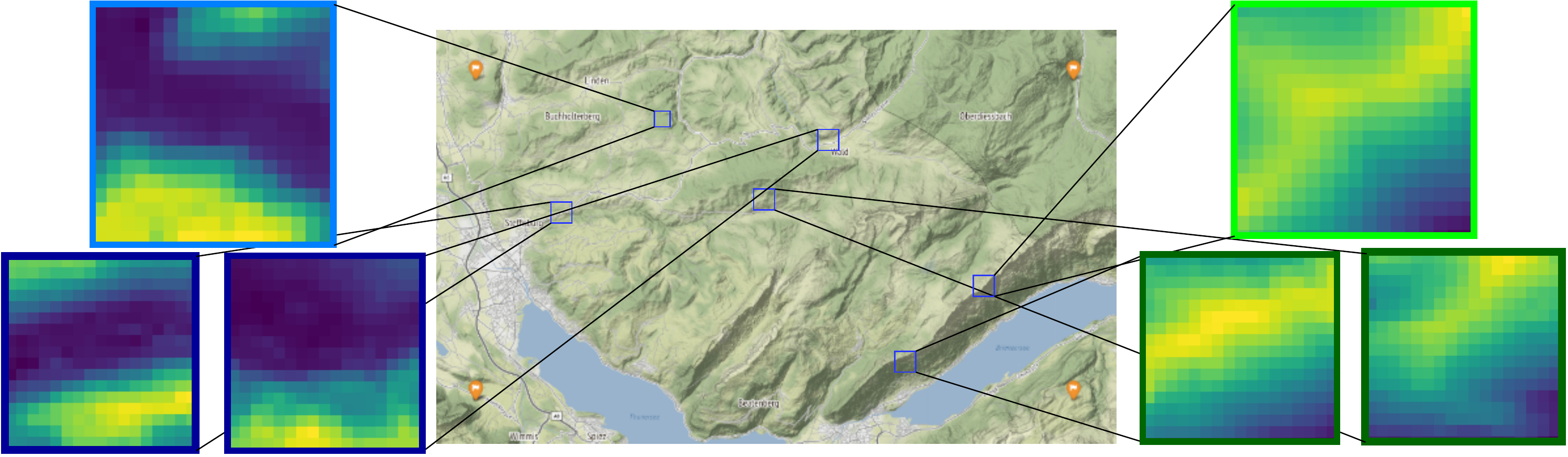} 
 	
    \caption{The top image of each side is the input location. The two bottom images are the two nearest neighbors of that input image in the latent space. This experiment was conducted in the orange rectangle, around $46^\circ46'30.0"N 7^\circ51'00.0"E$.}
    \label{fig:qual_knn}
\end{figure*}

We conducted another qualitative experiment to demonstrate the strength of our embedding space.
We took $n$ locations representing a topography pattern as input and averaged over their \ttv-4 embedding representations. 
We then used the KNN algorithm to find the ``most similar'' points in the embedding space to those input points. 
In Figure \ref{fig:qual_knn} we can see a representative example. 
We can see that our retrieved coordinates have very similar topography-patterns to those sampled.

This interactive experiment is also provided in the code.

\section{Related Work}
\label{sec:rw}
\subsection{Algorithmic Topography}

There have been works dealing with topography data enhancement and generation: 
Extracting information from raw elevation data \cite[]{JAEDICKE201431}, 
DTM (digital-terrain-model) extraction from satellite imagery \cite[]{GEVAERT2018106} and enhancing the resolution of topographic images \cite[]{Yue2015FusionOM}.
Many possible usages of DTM have been proposed. 
natural disasters analysis and prediction (such as Avalanche warning \cite[]{JAEDICKE201431, CHOUBIN2019123929} and
landslide susceptibility \cite[]{WU2009190}) to high solar energy regions localization \cite[]{HEO2020114588} and more \cite[]{cnn4glacier2020}.
In addition, deep learning has been used for the automatic mapping of topographic objects from DTM \cite[]{Torres2018ADL, TorresAlg4peaks2019}, and satellite imagery \cite[]{Li2020AutomatedTF}.

\subsection{Fractal-Effect in Topography}
The fact the earth's topography is a fractal has inspired some previous works. 
\cite{CHASE199239} studied the evolution of mountains and regional topography, and the effects of tectonic movement and climate on the landscape, including its fractal geometry.
\cite{Pelletier1997WhyIT} built a mathematical framework for explaining the existence of fractals in topography.
\cite{Weissel1995} used the fractal-effect in a research of the erosional development of the Ethiopian plateau of Northeast Africa and \cite{Liu2019AnIA} used it in a similar way for landslide susceptibility mapping.

\subsection{Unsupervised Learning for Visual and Geographic Data}

The concept of embedding an input to a latent lower-dimensional space is a basic and powerful one in the field of machine learning. As mentioned in the introduction, neural-networks tend to generate these spaces naturally during training.
Unsupervised learning for visual tasks is a very active field of research, making it very difficult to summarize. 
Recent methods that have pushed the state-of-the-are include: SimCLR \cite[]{Chen2020ASF}, MoCo \cite[]{He2020MomentumCF} and BYoL \cite[]{Grill2020BootstrapYO}.
All these methods follow the basic idea of making representation of an image similar under small transformations, exploiting several invariances within their domain such as invariance to rotation and color jitter. 
There have been some attempts in the geographic ML field to build specialized embedding spaces, most following Tobler's first law of geography \cite[]{tobler_geolaw}: ``\emph{everything is related to everything else, but near things are more related than distant things}''. 
Noteworthy examples are Tile2Vec \cite[]{Jean2019Tile2VecUR} which used the triplet-loss for spatially distributed data and the recent work of SauMoCo \cite[]{saumoco_2020} which takes this same basic idea from Tobler's law in order to create a new invariance for \textit{spatially close images}. We have seen in our experiments this assumption does not hold for classification on topographic classes.

\section{Conclusions}
\label{sec:conclusions}
In this work, we presented a novel self-supervision method for training neural networks for topographic feature extraction. We exploited the fractal-effect in the data during training and inference to build a model capable of generalizing to any relevant scale. 
Our key consideration in \ttv was leveraging the fractal-effect, which we achieved in an encoder-decoder based framework.
We evaluated our method on several topographic classes and topography-correlated classes and found it superior to other self-supervised methods and even surpassing fully supervised methods that used an order of magnitude more data.

Our results motivate a few interesting directions for future work we intend to explore. First is pushing empirical results further by using more elaborate architectures such as graph autoencoders \cite[]{kipf2016variational} and GANs \cite[]{Goodfellow2014GenerativeAN}.
Second, we plan to use our method for the automation of topography features mapping in multiple scales, greatly reducing the human labor necessary for such a task.

We hope this work will further advance the field of geographic ML, specifically to be used as additional features in tasks that exhibit correlation with the topography of its location (like avalanches and wildfire prediction). We also hope our method will act as a strong baseline for future works in the field of topography embedding.

\bibliography{topo2vec}

\end{document}